# Accelerating EM: An Empirical Study

Luis E. Ortiz* and Leslie Pack Kaelbling†
Computer Science Department, Box 1910
Brown University, Providence, RI 02912 USA
{leo,lpk}@cs.brown.edu

## Abstract

Many applications require that we learn the parameters of a model from data. EM (Expectation-Maximization) is a method for learning the parameters of probabilistic models with missing or hidden data. There are instances in which this method is slow to converge. Therefore, several accelerations have been proposed to improve the method. None of the proposed acceleration methods are theoretically dominant and experimental comparisons are lacking. In this paper, we present the different proposed accelerations and compare them experimentally. From the results of the experiments, we argue that some acceleration of EM is always possible, but that which acceleration is superior depends on properties of the problem.

## 1 INTRODUCTION

There are many applications in artificial intelligence and statistics that require the fitting of a parametric model to data. It is often desired to find the maximum-likelihood (ML) or maximum-*a-posteriori*-probability (MAP) model of the data. When all of the variables of the model are directly observable in the data, then this is relatively straightforward. When some variables are hidden, as is common in popular model classes such as Bayesian networks with hidden variables and hidden Markov models, maximum-likelihood parameter estimation is much more complicated.

The problem can be cast directly as an optimization problem: given a data set $D$ and a model of the form $\Lambda(\theta)$, find the setting of the parameters $\theta$ that maximizes the likelihood, $p(D|\Lambda(\theta))$ (or $p(\Lambda(\theta)|D)$ in the MAP case). Unfortunately, the objective function does not have a form that can be easily optimized globally, so we are generally reduced to local search methods.

* This material is based upon work supported in part under a National Science Foundation Graduate Fellowship and by DARPA/Rome Labs Planning Initiative grant F30602-95-1-0020.
† This work was supported in part by DARPA/Rome Labs Planning Initiative grant F30602-95-1-0020.

Probably the most naive strategy is to perform simple gradient ascent in the likelihood. This method has two problems. First, it is known to be relatively inefficient among the class of local search methods. Second, it is often the case that the model class $\Lambda$ constrains the choice of parameters $\theta$. For example, some of the components of $\theta$ may be constrained to describe a probability distribution, and so must be between 0 and 1 and must sum to 1. Naive gradient methods must be specially modified to respect such constraints.

The EM algorithm was described by Dempster et al. [1977] as a generalization of the Baum-Welsh algorithm for learning hidden Markov models [Rabiner, 1989]. Typically, it is monotonically convergent to a local optimum in likelihood space for probabilistic models, while directly satisfying possible constraints on the parameters.

There are two main classes of strategies for finding maximum-likelihood models with hidden parameters: accelerated gradient methods with constraint handling and EM. It has been observed empirically that, in some settings one algorithm appears to work better, and in other settings, other algorithms appear to work better. Furthermore, a number of researchers, both in the statistics and AI literatures, have proposed extensions, accelerations, and combinations of these methods.

In this paper, we seek to understand the relative merits of these optimization strategies and their extensions. Although there have been some attempts at theoretical comparisons of convergence rate [Xu and Jordan, 1996], the results are never clear-cut because they depend on properties of the individual problems to which they are applied. We have undertaken an empirical study in one of the simplest hidden-variable models: density estimation with a mixture of Gaussians. Given the infinite number of possible situations we can encounter, our only hope is to present empirical results suggesting the merits and disadvantages of each method in some situations. We believe that this study yields some insight into general properties of the methods, though the results cannot be guaranteed to transfer.

In the following sections, we describe the problem, some of the most common optimization methods available to solve it, the experiments we ran and the results we obtained. Please refer to the technical report [Ortiz and Kaelbling, 1999] for additional details.

## 2 DENSITY ESTIMATION WITH A MIXTURE OF GAUSSIANS

One of the simplest ML estimation problems with hidden variables is to model the probability density of a data set with a mixture of Gaussians. The model assumes that there is some number, $M$, of underlying "centers" in the $d$-dimensional space. Each data point is independently generated by first choosing a center with probability $\alpha_j$, and then drawing from a Gaussian distribution, with mean $\mu_j$ (the center) and covariance matrix $\Sigma_j$.

AutoClass [Cheeseman et al., 1988] casts the problem of how many centers to use in a Bayesian perspective by stating we should use the most probable number of centers given the data. In this paper, we address a subproblem of AutoClass: given a desired number of centers, find the model that maximizes the posterior probability.

In this problem, the parameter vector $\theta$ is made up of the $\alpha_j$, $\mu_j$ and $\Sigma_j$ for each $j$. From the independence of the data points, we can write the logarithm of the likelihood of the parameters with respect to the data, $p(D|\Lambda(\theta))$ as the sum of the logarithm of the likelihood with respect to the individual points,

$$L(\theta) = \ln p(D|\Lambda(\theta)) = \sum_{i=1}^{N} \ln p(x_i|\Lambda(\theta)), \quad (1)$$

where

$$p(x_i|\Lambda(\theta)) = \sum_{j=1}^{M} \alpha_j g(x_i|\mu_j, \Sigma_j), \quad (2)$$

and $g(x|\mu, \Sigma)$ is the multivariate Gaussian density with parameters $\mu$ and $\Sigma$. The constraints on $\theta$ are: (1) for all $j$, $\alpha_j > 0$; (2) $\sum_{j=1}^{M} \alpha_j = 1$; (3) for all $j$, $\Sigma_j$ is a symmetric, positive-definite matrix. It is sufficient to maximize $L(\theta)$ in order to maximize $p(D|\Lambda(\theta))$. Unfortunately, there is no direct method for performing ML estimation of $\theta$.

## 3 GRADIENT METHODS

The simplest gradient method we can use to find the maximum of the log-likelihood function is gradient ascent [Shewchuk, 1994, Bertsekas, 1995, Polak, 1971]. In this case, starting at some initial values for the parameters, at each iteration $k$, we obtain new values as follows:

$$\theta^{(k+1)} \leftarrow \theta^{(k)} + \gamma^{(k)} \nabla L(\theta^{(k)}), \quad (3)$$

where $\nabla L(\theta^{(k)})$ is the gradient of the log-likelihood function evaluated at the current values of the parameters and $\gamma^{(k)}$ is the step size we take uphill along the gradient direction. The value of $\gamma^{(k)}$ can be fixed or predetermined to decrease at every iteration.

This method does not respect any constraints on the parameters. To satisfy the constraints on $\alpha_j$, we project the part of the gradient relevant to these parameters into the constraint space and take a step size that does not take us out of this space [Bertsekas, 1995, Binder et al., 1997]. The projected gradient is

$$\nabla_\alpha^p L(\alpha^{(k)}) = \nabla_\alpha L(\alpha^{(k)}) - \frac{1}{M} \sum_{j=1}^{M} \left(\nabla_\alpha L(\alpha^{(k)})\right)_j, \quad (4)$$

where $(v)_j$ denotes the $j^{th}$ component of vector $v$.

Another strategy is to parameterize $\alpha$ in terms of another parameter $\omega$ such that $\omega$ is unconstrained and any assignment to $\omega$ satisfies the constraints on $\alpha$. One way of doing this is as follows:

$$\alpha_j(\omega) = \frac{e^{\omega_j}}{\sum_{j=1}^{M} e^{\omega_j}}. \quad (5)$$

To satisfy the constraint on $\Sigma_j$, we verify that the step size does not take us out of the constraint space and decrease the step size if it does until we get a step size that does not take us out of the constraint space.

Taking a fixed or predetermined step size at each step can slow down convergence. Instead, we can optimize the step size at every step by trying to find the largest value of the function in the direction of the gradient at every step by means of a line search; that is, this is the same as in gradient ascent but with

$$\gamma^{(k)} \leftarrow \operatorname*{argmax}_{\gamma} L(\theta^{(k)} + \gamma \nabla L(\theta^{(k)}))).$$

This method seems appealing but it has a drawback in that if the Hessian of the function at the local optimum is ill-conditioned (i.e., the function is "elongated"), it exhibits a zig-zagging behavior that can significantly slow down convergence [Bertsekas, 1995].

The conjugate gradient method tries to eliminate the zig-zagging behavior of optimized-step-size gradient ascent by requiring that we optimize along conjugate directions at every step. Formally, starting with some initial setting of the parameters $\theta^{(0)}$, $r^{(0)} \leftarrow \nabla L(\theta^{(0)})$, and the first direction $d^{(0)} \leftarrow r^{(0)}$, at each iteration $k$, we do as follows:

$\gamma^{(k)} \leftarrow \operatorname{argmax}_\gamma L(\theta^{(k)} + \gamma d^{(k)})$    *line search*
    *take best step in current direction*
$\theta^{(k+1)} \leftarrow \theta^{(k)} + \gamma^{(k)} d^{(k)}$
$r^{(k+1)} \leftarrow \nabla L(\theta^{(k+1)})$    *new gradient*
**if** $((k+1) \bmod d) == 0$ **then**
    $\beta^{(k+1)} \leftarrow 0$    *start over*
**else**    *weight to current direction*
    $\beta^{(k+1)} \leftarrow \frac{\left(r^{(k+1)}\right)^T (r^{(k+1)} - r^{(k)})}{\left(r^{(k)}\right)^T r^{(k)}}$
**end if**
    *new conjugate direction*
$d^{(k+1)} \leftarrow r^{(k+1)} + \beta^{(k+1)} d^{(k)}$

If we disregard the line search iterations, the conjugate gradient method has the appealing property that the number of iterations to convergence when *close to a solution* [1] is roughly equal to the number of parameters.

There are other more sophisticated methods, such as Newton and Quasi-Newton or variable metric, that try to solve the deficiencies of fixed- and optimized-step-size gradient ascent by reshaping the function. These methods use additional information about the function, like higher-order derivatives. However, using higher-order derivatives requires us to use additional storage and perform additional computations, which typically outweigh the reduction in the number of iterations.

## 4 EM BASICS

EM is a method for optimizing log-likelihood functions in the case of missing data or hidden variables [Dempster et al., 1977, McLachlan and Krishnan, 1997]. Starting with some initial value for the parameters, at each iteration, it uses the value of the parameters to compute the distribution or density over the hidden variables conditioned on the data (the expectation step) and then uses that distribution to get new values for the parameters (the maximization step). The likelihood function monotonically increases at each iteration, and under some regularity conditions on the likelihood function, the improvement is strict except at a stationary point of the likelihood function [Wu, 1983].

In the case of the mixture of Gaussians, at each iteration, we find, for each data point and each center, the conditional probability that the center generated the data point and then use that probability distribution to assign new values to the probability, mean and covariance matrix of each center. The expectation step is providing the maximization step with the information that will allow it to compute the expected sufficient statistics for each center under the conditional probability distribution over the center given the data and the current value of the parameters. EM uses the expected sufficient statistics in the maximization step as if they were the true sufficient statistics to obtain new values for the parameters.

More specifically, starting from some initial setting of the parameters, at each iteration $k$, the EM algorithm for mixture of Gaussians is

*E-step: compute distribution induced by $\theta^{(k)}$ and $D$*
**for** $i \leftarrow 1$ to $N, j \leftarrow 1$ to $M$ **do**
$$h_{ij}^{(k)} \leftarrow P(C = j|x_i, \Lambda(\theta^{(k)})) \propto \alpha_j^{(k)} g(x_i|\mu_j^{(k)}, \Sigma_j^{(k)})$$
**end for**
*M-step: update parameters*
**for** $j \leftarrow 1$ to $M$ **do**
$$\alpha_j^{(k+1)} \leftarrow \frac{\sum_{j=1}^{N} h_{ij}^{(k)}}{N}$$
$$\mu_j^{(k+1)} \leftarrow \frac{\sum_{j=1}^{N} h_{ij}^{(k)} x_i}{\sum_{j=1}^{N} h_{ij}^{(k)}}$$

---

[1] More formally, *close to a solution* means the neighborhood around the local optimum that can be well approximated by a quadratic function.

$$\Sigma_j^{(k+1)} \leftarrow \frac{\sum_{j=1}^{N} h_{ij}^{(k)} x_i x_i^T}{\sum_{j=1}^{N} h_{ij}^{(k)}} - \mu_j^{(k+1)} \left(\mu_j^{(k+1)}\right)^T$$
**end for**

For the mixture-of-Gaussians model, the method typically converges to a local maximum of the likelihood function, but it can stop in other stationary points, or it can go to a singular point where the likelihood function grows without bound [Redner and Walker, 1984]. A singular point in the mixture model occurs when we use data points as one of the means and let the variance of that center go to zero [Duda and Hart, 1973]. In practice, we can avoid singularities through good initializations. In cases where this is not enough, we can either (1) assign a small value to variances when they go below some small threshold value, (2) delete components with too-small variances, (3) use priors on $\Sigma_j$ and perform MAP estimation, or (4) restart EM with a different initial value for the parameters.

We can also view EM as a special form of the general gradient method. This allows us to see how EM reshapes the function it is optimizing to make it better conditioned [Xu and Jordan, 1996]. It also allows us to analyze its convergence theoretically. However, in the mixture-of-Gaussians model, the method gives updates that automatically satisfy the constraints on the parameters (in the case of $\Sigma_j$, this is true with probability 1 for sufficiently large $N$ [Redner and Walker, 1984, Xu and Jordan, 1996]). In practice, numerical errors can still take us out of the parameter space. In those cases, we can either: (1) take a smaller step in the direction of the EM update so as to guarantee constraint satisfaction, (2) eliminate components for which $\alpha_j = 0$ and put small thresholds on $\Sigma_j$, or (3) use priors on $\theta$ and perform MAP estimation.

## 5 ACCELERATION METHODS

Although EM has a very appealing monotonicity property, its convergence rate is significantly slow in some instances. For mixture of Gaussians, EM slows down when the centers (i.e., the Gaussian components) are very close together [Redner and Walker, 1984, Xu and Jordan, 1996].

One alternative is to start the optimization with EM and move to gradient methods when we are close to a solution. Another is to accelerate EM directly by using information about the EM iteration [McLachlan and Krishnan, 1997].

One of the direct accelerations is *parameterized EM* [Peters and Walker, 1978, Redner and Walker, 1984, Meilijson, 1989, Bauer et al., 1997]. Starting from some initial setting of the parameters, at each iteration, we get new values for the parameters as follows:

$$\theta^{(k+1)} \leftarrow \theta^{(k)} + \gamma^{(k)} \left(\theta_{EM}^{(k+1)} - \theta^{(k)}\right), \quad (6)$$

where $\theta_{EM}^{(k+1)}$ is the EM update with respect to the current parameters $\theta^{(k)}$. In this method, we use the change in values in the parameters at an EM iteration and take a step uphill in this direction from the current position or value of the parameters. The step size can be fixed as in gradient ascent, or optimized at every step as in optimized-step-size gradient ascent. It is equivalent to EM when $\gamma^{(k)} = 1$. In the case of mixture of Gaussians, we can achieve convergence using this method when we are *close to a solution* and $0 < \gamma^{(k)} < 2$ [Redner and Walker, 1984], and improve convergence speed when $\gamma^{(k)} > 0.5$ [Xu, 1997].

Parameterized EM is actually gradient or steepest ascent to find the zero of a function that is the change in parameters provided by EM (i.e., finding a fixpoint of the EM update).

Another acceleration method is conjugate gradient acceleration of EM [Jamshidian and Jennrich, 1993, Thiesson, 1995]. The idea is to use the change in value of the parameters of the EM iteration to find better conjugate directions when performing conjugate gradient. The method uses the information provided by the EM iteration to reshape the function and improve convergence speed when *close to a solution*. Formally, starting with some initial setting of the parameters $\theta^{(0)}$, $r^{(0)} \leftarrow \nabla L(\theta^{(0)})$, and the first direction $d^{(0)} = \theta_{EM}^{(1)} - \theta^{(0)}$, at each iteration $k$, we do as follows:

$\gamma^{(k)} \leftarrow \mathrm{argmax}_\gamma L(\theta^{(k)} + \gamma d^{(k)})$  *line search*
  *take best step in current direction*
$\theta^{(k+1)} \leftarrow \theta^{(k)} + \gamma^{(k)} d^{(k)}$
$r^{(k+1)} \leftarrow \nabla L(\theta^{(k+1)})$  *new gradient*
$u^{(k+1)} \leftarrow \theta_{EM}^{(k+1)} - \theta^{(k)}$  *EM direction*
**if** $((k + 1) \mod d) == 0$ **then**
  $\beta^{(k+1)} \leftarrow 0$  *start over*
**else**  *weight to current direction using EM information*
  $\beta^{(k+1)} \leftarrow -\frac{(u^{(k+1)})^T (r^{(k+1)} - r^{(k)})}{(d^{(k)})^T (r^{(k+1)} - r^{(k)})}$
**end if**
  *new conjugate direction*
$d^{(k+1)} \leftarrow u^{(k+1)} + \beta^{(k+1)} d^{(k)}$

This method is actually a special form of a generalized conjugate gradient method. The interesting aspect of the conjugate gradient acceleration of EM is that, contrary to the traditional generalized conjugate gradient method, it does not require the specification of a preconditioning matrix or the evaluation of second order derivatives or matrix-vector multiplications, since the change in parameters provided by the EM update rule approximates the generalized gradient.

We conjecture that the relationship between parameterized EM and conjugate gradient using EM information is similar to the relationship between fixed- and optimized-step-size gradient ascent and regular conjugate gradient: finding a good step size is problem dependent but optimizing the step size might not be a good idea (i.e., produces zig-zagging behavior) and moving in conjugate directions is better.

As pointed out by an anonymous reviewer, there are other extensions of the EM algorithm that can speed up convergence. Due to lack of space, we refer the reader to McLachlan and Krishnan [1997] for additional information about extensions and variations of EM. Two extensions of the EM algorithm are the *ECM (Expectation-Conditional Maximization)* [Meng and Rubin, 1993] and the *ECME (Expectation-Conditional Maximization Either)* [Liu and Rubin, 1994] algorithms. Both are useful when the regular maximization step is complex (i.e., no closed-form optimization) yet simpler if conditioned on a function of the current value of the parameters. Furthermore, ECME differs from ECM only in that it *conditionally maximize the log-likelihood function directly* in some of the steps. ECM typically has a slower convergence rate than EM, although it can be faster in actual total computing time. The convergence rate of ECME is typically faster than that of EM and ECM (the actual computing time to convergence is also typically faster). Given that the maximization step is simple in the context of the mixture-of-Gaussians model, these extensions do not really apply here except in the case that we reparameterized $\alpha$ as in equation (5). In this case, a version of an ECM algorithm can speed up convergence with respect to the typical alternative, which is an algorithm based on the generalized version of EM. However, it will still be typically slower than regular EM. We note that a version of the ECM algorithm can help learning in the context of the mixture-of-experts architecture [Jordan and Xu, 1993].

## 6 EXPERIMENTS

The theoretical convergence speed of the different methods presented is problem dependent. No one is theoretically dominant. There is empirical evidence that EM is superior to gradient ascent in the mixture-of-Gaussians case [Xu and Jordan, 1996]. The conjugate gradient and the acceleration methods work well when they are *close to a solution*. We argue that running conjugate gradient by itself is not a good idea since it requires a very precise line search when it is far from a solution. In practice, precise line searches can increase the time to convergence. Hence, the methods we compare to EM in this paper are all based on an idea that uses the monotonic convergence properties of EM. The algorithmic description is as follows [Jamshidian and Jennrich, 1993, Thiesson, 1995]: starting from some initial setting of the parameters,

**repeat**
  Run EM to get us *close to a solution*
  Run acceleration
**until** stopping condition

When needed, we use inexact line searches during the accelerations to save time. This seems to work well when we are close to a solution. However, a decrease in log-likelihood can occur due to the inexact line search. If a decrease in log-likelihood occurs during the acceleration, we return to EM and repeat the process. We interpret the condition *close to a solution* to be true when the change in log-likelihood is less than 0.5. This means that we continue to run EM "as long as the $\chi^2$ statistic for testing the equality of two successive iterates is more than 1" [Jamshidian and Jennrich, 1993].

When needed, the line search we use is an adapted version of the secant-like line search used by Jamshidian and Jennrich [1993]. Note that even in methods that do not use a line search, like parameterized EM, we still need to make sure that the step size does not take us outside the constraint space. In those cases, we reduce the step size until we find one that keeps us inside the parameter space.

The basis of our empirical analysis is the idea that the work needed to compute both the gradient and EM update is approximately the same. Actually, as long as $N$ is sufficiently large such that the extra computation is not significant, we can compute both in about the same time, since they require about the same information. We can see this from the expression of the gradient [Xu and Jordan, 1996]. Let the expected counts for center $j$ be $N_j = \sum_{i=1}^{N} h_{ij}$, and let $\alpha_j^{EM}$, $\mu_j^{EM}$ and $\Sigma_j^{EM}$ be the result of applying the EM update rule to $\alpha_j$, $\mu_j$ and $\Sigma_j$, then

$$\frac{\partial L(\theta)}{\partial \alpha_j} = \frac{N_j}{\alpha_j} = N\frac{\alpha_j^{EM}}{\alpha_j}, \quad (7)$$

$$\nabla_{\mu_j} L(\theta) = N_j \Sigma_j^{-1}(\mu_j^{EM} - \mu_j), \quad (8)$$

$$\nabla_{\Sigma_j} L(\theta) = \frac{1}{2} N_j \text{vec}\left(\Sigma_j^{-1}\left(\Sigma_j^{EM} - \Sigma_j + (\mu_j^{EM} - \mu_j)(\mu_j^{EM} - \mu_j)^T\right)\Sigma_j^{-1}\right). \quad (9)$$

First of all, we note that both require the computation of the expected sufficient statistics $\sum_{i=1}^{N} h_{ij}$, $\sum_{i=1}^{N} h_{ij} x_i$, and $\sum_{i=1}^{N} h_{ij} x_i x_i^T$ to obtain the EM update. This takes $O(NMd^3)$ ($O(NMd)$ when dealing with independent features) [2]. Using the EM update, computing the EM direction takes $O(Md^2)$ extra work and computing the gradient takes $O(Md^3)$ extra work (in case of independent features, both bounds are $O(Md)$). We also note that the constants in the bounds for extra work are small. Therefore, for sufficiently large $N$, the time to compute the expected sufficient statistics dominates all others. Finally, we say that the computation of the gradient, the EM iteration, and both at the same time are all *EM-equivalent* iterations. This way we do not need to compare CPU times, which significantly depend on the implementation details of the different methods. All we need to do is optimize the methods with respect to EM-equivalent iterations.

There are many different initialization methods. Some of them have been studied for large-dimensional data in the context of the naive Bayesian Network model [Meilă and Heckerman, 1998]. For simplicity, we use the following initialization:

Initialize $\alpha$ as a uniform random sample from the space of all distribution over $M$ events.

Initialize $\mu$ for each center by sampling uniformly at random from the space defined by the hypercube of minimum volume containing all the data points.

Initialize $\Sigma$ for each center as a diagonal matrix with variances equal to the square of the distance to the center closest to it [Bishop, 1996].

---

[2]This is assuming that the complexity of exponentiation is $O(1)$.

One remaining issue is when to stop. Ideally, an iterative method should stop when it has reached values for the parameters such that they provide a "good" model. However, there is no clear way for an iterative method to determine this. Therefore, detecting when to stop is a crucial but hard problem in general. For instance, it is common to encounter situations where the function we want to optimize has many areas of large and small changes towards the local optimum. In turn, this causes some of the methods to produce burst of large and small improvements, which must be handled by stopping rules. Many different stopping rules have been used in the optimization literature [Bertsekas, 1995]. In this paper, we do not deal with the stopping problem and use a very simple, typical stopping rule based on the progress of the method in log-likelihood (or log-posterior) space [3]. We stop when the change in log-likelihood from one iteration to the next is less than $10^{-5}$. We can obtain the information we need to test this condition easily from the EM-equivalent iteration. In our experiments on synthetic data, all the methods that we tested converged to the same point in log-likelihood space (and parameter space, sometimes modulo symmetrically equivalent models [4]).

All the methods we tested, besides regular EM (**EM**), have the algorithmic structure presented above and only the acceleration step differed. We tried the following accelerations:

- regular conjugate gradient (**CG**),
- conjugate gradient with EM information (**CG+EM**),
- parameterized EM with inexact line search to optimize the step size (**PEM(opt)**),
- parameterized EM with fixed step sizes 1.5 (**PEM(1.5)**) and 1.9 (**PEM(1.9)**).
- conjugate gradient with EM information on reparameterized ($\omega$) space [5] (**CG+EM(rp)**).

In order to examine the properties of the different methods, we tested them on data we generated from simple models with varying degree of separation between the Gaussians. We generated data from 3 models with 2 Gaussians in 2-dimensions. All 3 models had the same parameters $\alpha$ and $\Sigma$ ($\alpha_1 = \alpha_2 = 0.5, \Sigma_1 = \Sigma_2 = I$). The first center $\mu_1$ for one of the Gaussians in the 3 models was also the same ($\mu_1 = (0, 0)$). The models differed in the center of the second Gaussian: (1) $\mu_2 = (3, 3)$, (2) $\mu_2 = (2, 2)$, and (3) $\mu_2 = (1, 1)$. We generated one data set of 2000 points from each of the models (See Figure 1). We then generated 40 initial sets of parameters using the method presented above and ran each algorithm on each data set starting from each one of those initial parameters. Table 1 presents the results. For each data set, the results in the first column are the average number of EM-equivalent iterations for each method. The results in the second column are the average speed-up. We define the speed-up achieved by a proposed acceleration method in a run as the number of iterations of EM divided by the number of EM-equivalent iterations of the method for that run. Average speed-up may be a better measure because it is not so drastically influenced by cases that are hard for everyone.

---

[3]In theory, convergence speed in log-likelihood space is faster than in parameter space.

[4]In almost all cases, the methods converged to the same parameter values (i.e., equivalent models) and therefore there was no need to compare them in terms of KL-divergence of the resulting model from the true model.

[5]The EM algorithm in this case is actually a Generalized EM algorithm since there is no exact (i.e., closed-form) optimization in the maximization step.

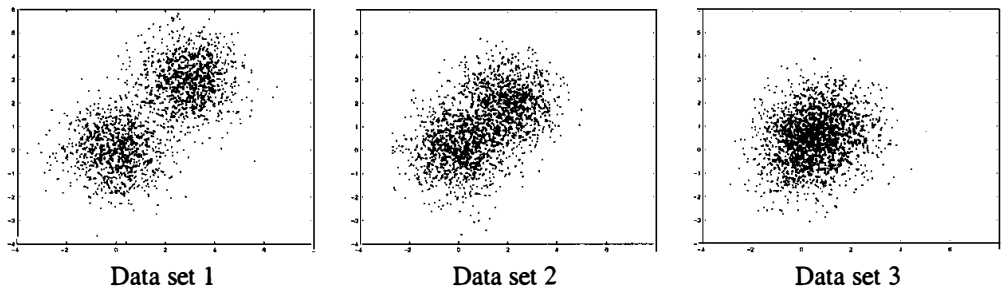

Figure 1: Data sets 1, 2, and 3.

| Method | Data Set | | | | | |
|---|---|---|---|---|---|---|
| | 1 : (0,0)-(3,3) | | 2 : (0,0)-(2,2) | | 3 : (0,0)-(1,1) | |
| | num. iters. | speed-up | | | | |
| EM | 120 | | 198 | | 2356 | |
| CG | 163 | 0.78 ± 0.07 | 225 | 1.04 ± 0.07 | 585 | 3.98 ± 0.38 |
| CG+EM | 100 | 1.18 ± 0.19 | **122** | **1.78 ± 0.19** | 187 | **12.80 ± 1.50** |
| CG+EM(rp) | 116 | 1.04 ± 0.18 | 131 | 1.70 ± 0.19 | 214 | 11.92 ± 1.46 |
| PEM(opt) | 120 | 1.01 ± 0.10 | 202 | 1.02 ± 0.05 | 1967 | 1.58 ± 0.18 |
| PEM(1.5) | **83** | **1.40 ± 0.04** | 137 | 1.44 ± 0.02 | 1642 | 1.41 ± 0.02 |
| PEM(1.9) | 81 | 1.32 ± 0.09 | 110 | 1.79 ± 0.03 | 1329 | 1.74 ± 0.04 |

Table 1: This table presents the average number of EM-equivalent iterations that each method took to converge on the different data sets for the first set of experiments. Also in the table are the (approximate) 95% confidence intervals on the average speed-up of the methods with respect to the number of iterations taken by EM (i.e., speed-up of a run = number of iterations of EM / number of iterations of acceleration for that run).

We performed a bootstrap version (shift method) of the one-sided paired-sample test [Cohen, 1995] to compare the method with the best average number of iterations and/or speed-up with each of the other methods. For each data set, results are in bold for the method with the best empirical mean. The test did not reject the null hypothesis that the difference in mean was significant ($p \leq 0.05$, $K = 10000$) with respect to the method with the best empirical mean only for the other methods with the results in bold.

The results from this experiment show that accelerating with **CG+EM** and **CG+EM(rp)** significantly improve convergence speed as the Gaussians get closer together. When the Gaussians are farther apart, all accelerations except **PEM(1.5)** and **PEM(1.9)** can slow down convergence due to false starts of the acceleration (false signalings of closeness to a solution). However, the improvement in conver-gence speed provided by **PEM(1.5)** and **PEM(1.9)** is not as impressive as that of **CG+EM** and **CG+EM(rp)** in hard instances. Also, the slow-downs produced by the attempted accelerations tend to occur mainly in easier instances (models with means at (0,0)-(3,3) and (0,0)-(2,2)) of the problem and are not as severe as the potential improvements in hard instances (model with means at (0,0)-(1,1)). Figure 2 presents a scatter plot of the behavior of **CG+EM** in data set 3. Finally, note from Figure 3 and the small values of the confidence intervals for the speed-up for the fixed-step-size parameterized EM method that the behavior of fixed-step-size parameterized EM methods seems in general very consistent.

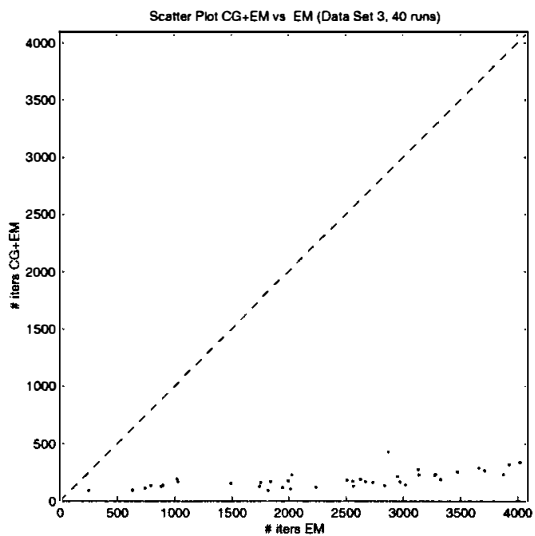

Figure 2: Scatter plot of the number of EM-equivalent iterations of **CG+EM** vs. the number iterations of EM on data set 3.

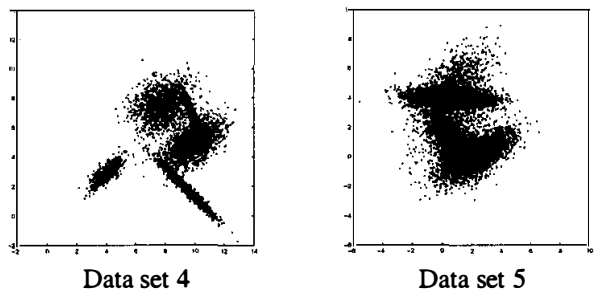

Figure 4: Data set 4 and 5.

| Method | Data Set | | | | | |
|---|---|---|---|---|---|---|
| | 6 : ($M = 5, d = 5$) | | 4 : ($M = 5, d = 2$ (easy)) | | 5 : ($M = 5, d = 2$ (hard)) | |
| | num. iters. | speed-up | num. iters. | speed-up | num. iters. | speed-up |
| EM | 122 | | 136 | | 1672 | |
| CG+EM | 140 | $1.20 \pm 0.32$ | 140 | $1.05 \pm 0.25$ | 195 | $7.98 \pm 1.36$ |
| PEM(opt) | 140 | $1.01 \pm 0.10$ | 121 | $1.14 \pm 0.50$ | 1028 | $2.02 \pm 0.28$ |
| PEM(1.5) | 91 | $1.13 \pm 0.04$ | 111 | $1.12 \pm 0.06$ | 1160 | $1.42 \pm 0.03$ |
| PEM(1.9) | 89 | $1.13 \pm 0.06$ | 100 | $1.20 \pm 0.08$ | 940 | $1.73 \pm 0.05$ |

Table 2: This table presents the average number of EM-equivalent iterations that each method took to converge and the (approximate) 95% confidence intervals on the average speed-up on the different data sets for the second set of experiments.

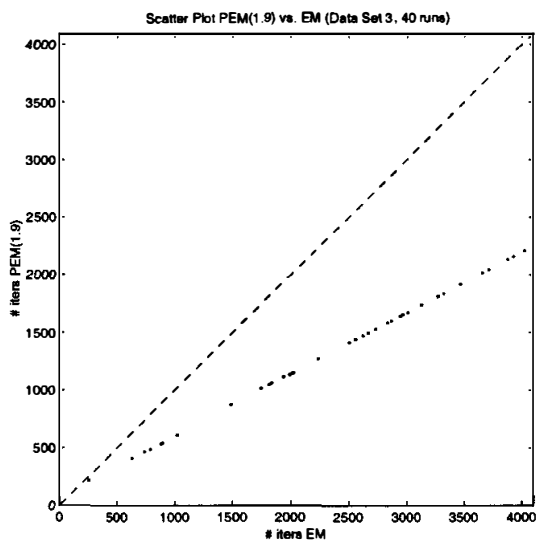

Figure 3: Scatter plot of the number of EM-equivalent iterations of **PEM(1.9)** vs. the number of EM iterations on data set 3. This linear behavior is typical of fixed-step-size parameterized EM on all the data sets.

Although the results are not reported in this paper, we also tried running EM on the reparameterized space alone and running conjugate gradient alone in all the experiments but they had much less success on average.

We also ran experiments with models of more Gaussians and/or higher dimensions with similar results. Models 4, 5 and 6 are random models with different characteristics. Models 4 and 5 both have $d = 2$ and $M = 5$. They differ in that in the data generated from model 5 it is harder to distinguish the different clusters than in that generated from model 4 (See Figure 4). Model 6 is larger with $d = 5$, $M = 5$. We generated a data set of 5000 points from model 4, 10000 points from model 5, and 20000 points from model 6. Table 2 presents the results. The results are averages of 40, 40 and 38 random initial settings of the parameters for models 4, 5 and 6 respectively. Again, we have **CG+EM** being superior in the hard case, and while it might slow down in the easy cases, the slow down does not seem that severe. We note again that **PEM(1.5)** and

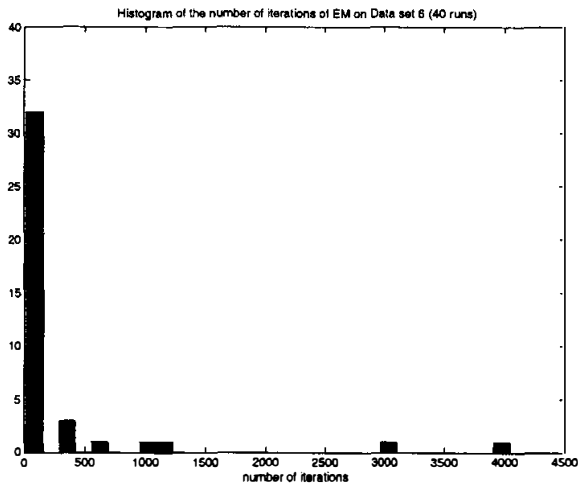

Figure 5: Histogram of the number of iteration of EM on data set 6.

| Method | IRAS data results | |
|---|---|---|
|  | num. iters. | speed-up |
| EM | 480 |  |
| CG+EM | 474 | 0.99 |
| **PEM(1.9)** | **272** | **1.66** |

Table 3: This table presents the average number of EM-equivalent iterations and speed-up that each method took to converge on the IRAS data set starting with initially $M = 77$.

**PEM(1.9)** seem almost consistently better than **EM**. Finally, we note that there are 2 runs missing from the results in the table for model 6. For one of those runs,**EM** took 4041 iterations on that run compared to 2974 for **PEM(1.5)**, 2483 for **PEM(1.9)** and 21456 for **PEM(opt)**. We stopped **CG+EM** when it had more iterations than all the others. **CG+EM** and **PEM(opt)** were failing during the line search because they were running out of time [6]. The value of the log-likelihood for the point where all the methods converged in that run was smaller than the most common one. We suspect that this is a saddle point or some flat region in log-likelihood space. For the other run, textbfEM took 2976 iterations on that run, compared to 180 for **CG+EM**, 5836 for **PEM(opt)**, 1999 for **PEM(1.5)**, and 1998 for **PEM(1.9)**. Inspection of the behavior of **PEM(opt)** showed that the method was wasting time because the line searches were failing almost immediately, and therefore going back to EM immediately after each attempt. For this run, however, the point where the methods converged was the most common point in log-likelihood space. Finally, we note that the behavior of both of these two runs was uncommon as suggested by the histogram of the number of iterations of **EM** for this data set in Figure 5; the two largest values in the histogram are for the runs mentioned above.

As an anonymous reviewer pointed out, the stopping rule we use does not have a scale. Therefore, it is insensitive to the range of the log-likelihood function. Restating the issue of stopping rules, we note that partial preliminary experiments suggest that a *scaled* stopping rule [7] can indeed help reduce the number of iterations required by **EM** and **PEM** in hard instances. In such cases, the log-likelihood function is ill-conditioned and stopping anywhere in the relatively flat region close to the solution produces very good estimates with regard to maximizing log-likelihood. It can also help in preventing over-fitting when the amount of data is small; a very important issue when we are learning models from data. However, partial preliminary experiments also suggest that, unless we use different threshold values for different instances, a scaled stopping rule increases the potential for stopping too early in easier instances, thus producing bad estimates. Other stopping rules have been used for EM, but we do not know of any study that has been done to compare them. Finally, we conjecture that the "right" stopping rule eliminates the need for most of the type of accelerations proposed since it eliminates the basis for them. This is because the proposed accelerations work well when we are close to a solution and the problem is ill-conditioned, which is exactly what the "right" stopping rule would detect.

We also ran experiments on the Infrared Astronomical Satellite (IRAS) data set [Cheeseman et al., 1988, Cheeseman and Stutz, 1995]. This data set contains $N = 5420$ data points in $d = 93$ dimensions. AutoClass found $M = 77$ classes using a mixture-of-Gaussians model with independent features. Given the high dimensionality and other problems that this data poses, we performed MAP estimation over a model with independent features and took careful steps in the way we computed the expected sufficient statistics and the value of the change in log-posterior. Therefore, our implementation is similar to that of AutoClass.

We assumed *a priori* that the parameters were independent. We used a Dirichlet prior on the $\alpha$ parameters with the same prior counts which we set to $(M + 1)/M$. We used a uniform prior over the hypercube of minimum volume containing all the data points for the mean parameters of the Gaussian components. We used a (scaled) inverse-$\chi^2$ prior for the variances with $1/M$ degrees of freedom and scale 1. In some instances, the update rule took us close to the boundary constraint. Our solution to this problem was to eliminate components if their probability was less than $1/N$ and/or at least one of their variances was less than $10^{-100}$. Once we removed those invalid components, we restarted the method using the value of the parameters of the remaining components as the initial value of the parameters.

---

[6]The inexact line search that we used had a optimum number of trials before failing which was set to 10 [Jamshidian and Jennrich, 1993].

[7]For instance, as suggested by that anonymous reviewer, we stop when the change in log-likelihood at one iteration relative to the total change so far is smaller than some threshold (i.e., $10^{-5}$).

We ran experiments assuming a mixture-of-Gaussians model with initially 77 components. We generated random initial parameters using an adapted version of the initialization procedure used by AutoClass C and ran each method starting from each one of the initial parameters. Table 3 presents the results based on 5 runs. First, we note that EM is very stable at the beginning and makes most of its progress in the first 20 − 25 iterations. This behavior has also been noted by others [Redner and Walker, 1984, Xu and Jordan, 1996]. Fixed-step-size parameterized EM tends to perform best in this data set with respect to speed of convergence. Conjugate gradient acceleration of EM can slow down convergence. To understand this result we note that the shape of the log-posterior seems to have many valleys, as suggested by the typical behavior of EM on this data as shown in Figure 6. Therefore, we make many false signalings of closeness and start the acceleration before it is really close to a solution. The methods that use line searches wasted time, since most of the line searches fail and therefore the attempt to accelerate fails. Again, this type of behavior suggests that we need better ways of signaling closeness or stopping. Preliminary analysis seems to indicate that this behavior of EM does not simply result from performing MAP estimation. Therefore, we wonder whether the cause of this is related to the fact that the ratio of the number of parameters to the number of samples is not small enough, or this is just a consequence of the high dimensionality, or both [8].

## 7 CONCLUSIONS

First of all, we note, as many others authors have before us, that the performance of EM away from the solution is impressive, particularly in log-likelihood (and log-posterior) space. Given the "right" stopping rule for a problem, it typically produces reasonable estimates relatively fast. In addition, it typically exhibits global convergence [9] in practice. These properties, along with its monotonicity property and its simplicity, make EM a very powerful method and a first choice for finding ML (or MAP) estimates in the context of the mixture-of-Gaussians and other probabilistic models.

Nevertheless, using the assumption that the amount of data is sufficiently large such that the extra computation of the gradient and the EM direction is not significant and a simple albeit conservative stopping rule, our experimental results on synthetic data suggest that the method based on conjugate gradient acceleration of EM can be a good choice for finding ML estimates for the mixture-of-Gaussians model. This is because it significantly improves convergence in the hard cases and, while it can slow down convergence in the easy cases, the slow-down is not as severe, given the relatively low number of iterations required to converge in those cases. In addition, although it is more complicated to implement than parameterized EM, it eliminates the setting of the step size parameter. Furthermore, although it requires line searches, those searches can be simple and inexact in a neighborhood close to a solution. Finally, the behavior of conjugate gradient acceleration of EM seems best when the function is smooth but very flat and "elongated" in the neighborhood of the local optimum.

---

[8]Numerical instability can certainly be a reason too.

[9]Here, *global convergence* means the property of an optimization method to be able to converge to a stationary point (not necessarily a global optimum) in function space from any initial point in the parameter space.

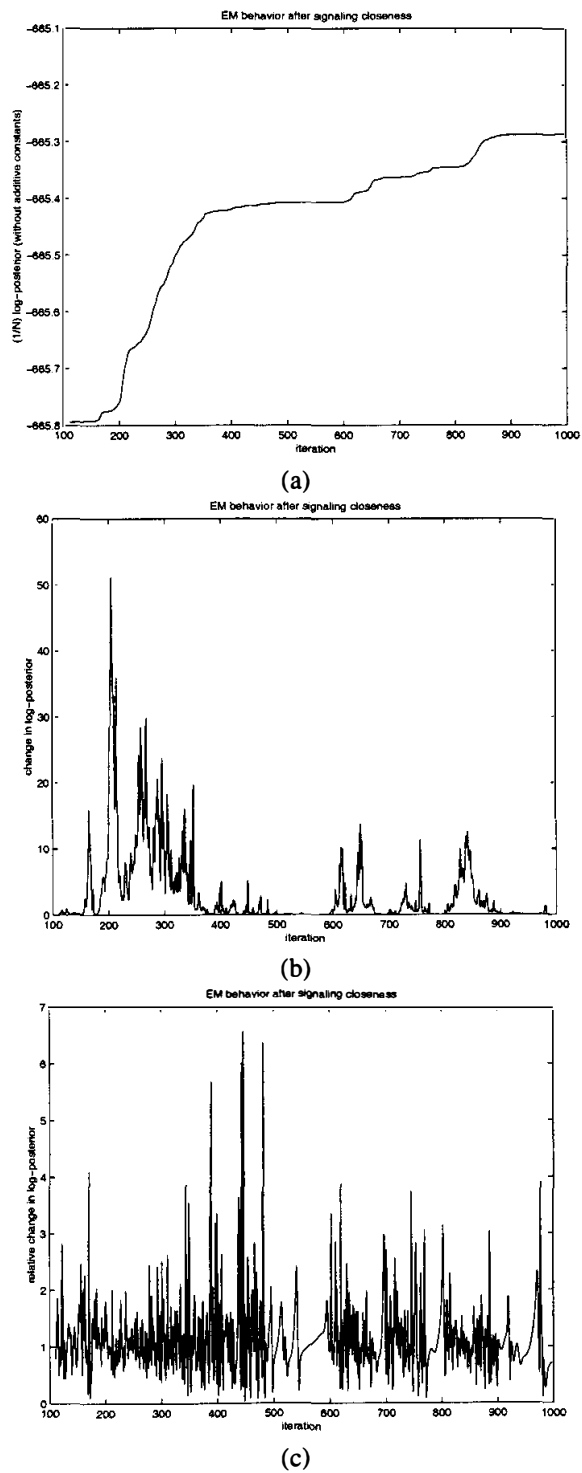

Figure 6: Plots of (a) the log-posterior (without constants) divided by $N$, (b) the change in log-posterior and (c) the relative change in log-posterior after signaling closeness for a typical run of EM on the IRAS data set. We plot the log-posterior/$N$ as opposed to the log-posterior itself to reduce the scale of the y-axis. The relative change is the current change divided by the previous change and can be used as an approximation of the convergence rate.

On the other hand, the results from the experiments on the IRAS data suggest that it is a good idea to attempt a simple acceleration method, such as fixed-step-size parameterized EM, before trying the more complicated conjugate-gradient-based accelerations. This is because there are cases in which the surface of the log-likelihood (or log-posterior) is relatively flat but not very smooth in the neighborhood of the local optimum. We can attempt those more complicated methods once we note that the simple acceleration method is still too slow.

We think that it is necessary to perform a similar comparison analysis in the context of learning Bayesian networks and HMMs to verify that the same characterization of superiority of the accelerations based on "easy" and "hard" instances suggested in this paper carries over.